\documentclass[10pt, conference]{IEEEtran}
\IEEEoverridecommandlockouts
\usepackage{cite}
\usepackage{amsmath, amssymb, amsfonts}
\usepackage{graphicx}
\usepackage{booktabs}
\usepackage{array}
\usepackage{caption}
\usepackage{placeins}
\usepackage{dblfloatfix}
\usepackage{url}
\usepackage{hyperref}

\def\BibTeX{{\rm B\kern-.05em{\sc i\kern-.025em b}\kern-.08em T\kern-.1667em\lower.7ex\hbox{E}\kern-.125emX}}

\begin{document}
    \title{EMCStereo: Attention-Enhanced Stereo Matching for Thin-Structure Depth
    Estimation with a Synthetic Tree-Branch Benchmark}

    \author{\IEEEauthorblockN{Yida Lin, Bing Xue, Mengjie Zhang} \IEEEauthorblockA{\small \textit{Centre for Data Science and Artificial Intelligence} \\ \textit{Victoria University of Wellington, Wellington, New Zealand}\\ linyida\texttt{@}myvuw.ac.nz, bing.xue\texttt{@}vuw.ac.nz, mengjie.zhang\texttt{@}vuw.ac.nz}
    \and \IEEEauthorblockN{Sam Schofield, Richard Green} \IEEEauthorblockA{\small \textit{Department of Computer Science and Software Engineering} \\ \textit{University of Canterbury, Canterbury, New Zealand}\\ sam.schofield\texttt{@}canterbury.ac.nz, richard.green\texttt{@}canterbury.ac.nz}}

    \maketitle
    \vspace{-1.4em}

    \begin{abstract}
        Thin structures such as tree branches are one of the hardest cases for
        stereo matching: a branch is only a few pixels wide, the background
        behind it is cluttered and weakly textured, and dense ground truth for
        real branches is almost impossible to label by hand. We make three
        contributions. \textbf{EMCStereo} places three lightweight attention
        modules inside a PSMNet-style cost-volume backbone: Efficient Multi-scale
        Attention (EMA) on the deep semantic feature, a Multi-Scale Fusion block
        (MSFblock) that learns weights over the spatial pyramid instead of
        concatenating it, and Coordinate Attention (CoordAtt) on the final
        matching feature. Because MSFblock collapses the four pyramid branches
        into one, the modules leave the network $2.0\%$ \emph{smaller} than the
        same backbone without them and take $1.7\%$ of its inference time.
        \textbf{VirtualTree} is a synthetic stereo dataset rendered in Unreal
        Engine~5 with a simulated ZED Mini rig: $5{,}520$ pairs whose EXR depth
        gives dense, exact disparity for thin-branch geometry. An
        \textbf{eight-way ablation} with an independent repeat of the
        no-attention baseline puts the run-to-run noise floor at $0.009$\,px
        end-point error (EPE), while re-evaluating one checkpoint on different
        hardware agrees to $4\times10^{-6}$\,px. EMCStereo reaches $1.31$\,px EPE
        at $5.96\%$ D1-all on the held-out VirtualTree test split, $1.00$\,px on
        a held-out SceneFlow split, and $0.80$, $0.73$, $0.62$ and $3.19$\,px on
        KITTI~2012, KITTI~2015, ETH3D and Middlebury, with $\delta_{1}$ depth
        accuracy from $92.6\%$ to $98.7\%$. Read against the noise floor, the
        attention stack is accuracy-neutral at a matched $100$-epoch budget, and
        MSFblock and CoordAtt cost $0.03$--$0.05$\,px unless EMA is present.
    \end{abstract}


    \vspace{-0.8em}
    \section{Introduction}

    Stereo matching recovers a dense disparity map, and hence metric depth, from
    a rectified image pair without active ranging hardware. That makes it
    attractive for field robotics in agriculture and forestry, where a robot has
    to perceive complex vegetation before it can navigate, inspect or handle a
    plant~\cite{lin2024branch, lin2025segmentation}. Robotic tree pruning is a
    demanding case: a manipulator or a UAV-mounted cutter must locate thin
    branches in three dimensions before it approaches
    them~\cite{lin2024branch, lin2025yolosgbm}.

    Thin structures are close to a worst case. A branch may be three or four
    pixels wide, it occludes itself and its neighbours, and bark carries little
    distinctive texture; foliage is semi-transparent and moves in the wind, and
    the sky behind a tree creates abrupt depth steps. Cost-volume networks smooth
    over these high-frequency regions, so the structures that matter most are
    blurred or lost. Supervision is the second problem: dense, pixel-accurate
    disparity for real branches is out of reach for both LiDAR and manual
    labelling.

    Deep stereo has moved quickly, from the cost-volume regression of
    GC-Net~\cite{kendall2017gcnet} and PSMNet~\cite{chang2018psmnet} to
    group-wise correlation~\cite{guo2019gwcnet}, guided
    aggregation~\cite{zhang2019ganet} and recurrent
    refinement~\cite{lipson2021raft}, while attention has become a standard way
    to sharpen dense-prediction
    features~\cite{hu2018senet,woo2018cbam,hou2021coordatt,ouyang2023ema}. Hence
    our question: can a few lightweight attention modules, placed at the right
    points of a proven cost-volume backbone, recover the thin-structure detail
    that plain aggregation loses? We answer it by measurement. Alongside an
    eight-way module grid we repeat the baseline run from scratch, which fixes
    the resolution of the comparison at $0.009$\,px EPE; against that floor the
    answer is a qualified no. At equal budget the modules buy a smaller and
    slightly cheaper model, not a lower error.

    For supervision we turn to simulation. Unreal Engine~5 (UE5)~\cite{ue5}
    renders photorealistic trees and exports exact depth for every pixel,
    including branches nobody could label by hand. We render
    \textbf{VirtualTree} with a simulated ZED Mini rig~\cite{zedmini} and convert
    its EXR depth to disparity through the calibrated stereo geometry. The
    network, \textbf{EMCStereo}, is named after the three modules it adds:
    \textbf{E}MA, \textbf{M}SFblock and \textbf{C}oordAtt. Our contributions are
    (i) \textbf{EMCStereo}, three complementary attention modules inserted into a
    single-cost-volume PSMNet pipeline while \emph{reducing} the parameter count;
    (ii) \textbf{VirtualTree}, $5{,}520$ simulated ZED Mini pairs with dense,
    exact ground truth for thin branches; and (iii) an \textbf{eight-way ablation
    with a measured noise floor}, which leaves one effect standing: MSFblock and
    CoordAtt cost $0.03$--$0.05$\,px unless EMA is present. We evaluate mainly on
    VirtualTree and also on SceneFlow~\cite{mayer2016large},
    KITTI~2012~\cite{geiger2012kitti}, KITTI~2015~\cite{menze2015kitti},
    ETH3D~\cite{schops2017eth3d} and Middlebury~\cite{scharstein2014middlebury}.

    \vspace{-0.85em}
    \section{Related Work}

    Learning-based stereo began with patch-matching costs~\cite{zbontar2016mccnn}
    and direct correlation regression~\cite{mayer2016large}, later refined by
    cascaded residual sub-networks~\cite{pang2017crl, liang2018iresnet} and
    semantic cues~\cite{yang2018segstereo}. End-to-end cost-volume networks then
    took over: GC-Net~\cite{kendall2017gcnet} introduced differentiable
    soft-argmin regression over a 3D volume, and PSMNet~\cite{chang2018psmnet}
    added spatial pyramid pooling and stacked hourglass 3D convolutions. Later
    work reworked the volume and its aggregation through group-wise
    correlation~\cite{guo2019gwcnet}, guided aggregation~\cite{zhang2019ganet},
    3D-convolution-free adaptive aggregation~\cite{xu2020aanet}, differentiable
    PatchMatch pruning~\cite{duggal2019deeppruner}, attention-filtered
    volumes~\cite{xu2022acvnet} and recurrent updates~\cite{lipson2021raft}, and
    foundation-model stereo such as DEFOM-Stereo~\cite{jiang2025defom} now
    injects monocular depth priors. EMCStereo keeps the compact
    single-cost-volume PSMNet template and asks how far targeted attention can
    push it.

    Attention re-weights features so that informative channels and locations
    dominate. SENet~\cite{hu2018senet} recalibrates channels,
    CBAM~\cite{woo2018cbam} adds a spatial branch,
    CoordAtt~\cite{hou2021coordatt} factorizes attention along the height and
    width axes to keep the positional structure global pooling discards,
    EMA~\cite{ouyang2023ema} groups channels and couples a cross-spatial branch
    with a $3{\times}3$ branch, and the MSFblock of
    SHISRCNet~\cite{chen2023shisrcnet} learns softmax weights over receptive
    fields. All four are light and architecture-agnostic, yet rarely combined or
    deliberately placed inside a stereo backbone.

    Synthetic rendering underpins stereo learning: the SceneFlow
    suite~\cite{mayer2016large} enabled the first end-to-end networks, and
    engines such as UE5~\cite{ue5} export exact per-pixel geometry. Vegetation is
    badly served by real datasets, because branches are narrow, self-occluding
    and impractical to label. Earlier forestry work tuned classical stereo with
    genetic algorithms~\cite{lin2025genetic}, paired YOLO detection with SGBM
    stereo to measure branch distance~\cite{lin2024branch, lin2025yolosgbm} and
    evaluated deep segmentation of branches~\cite{lin2025segmentation}; we use
    UE5 as a controllable source of exact thin-branch supervision.

    \vspace{-0.85em}
    \section{The EMCStereo Network}

    \subsection{Overview}

\begin{figure*}[!t]
        \centering
        \includegraphics[width=0.84\textwidth]{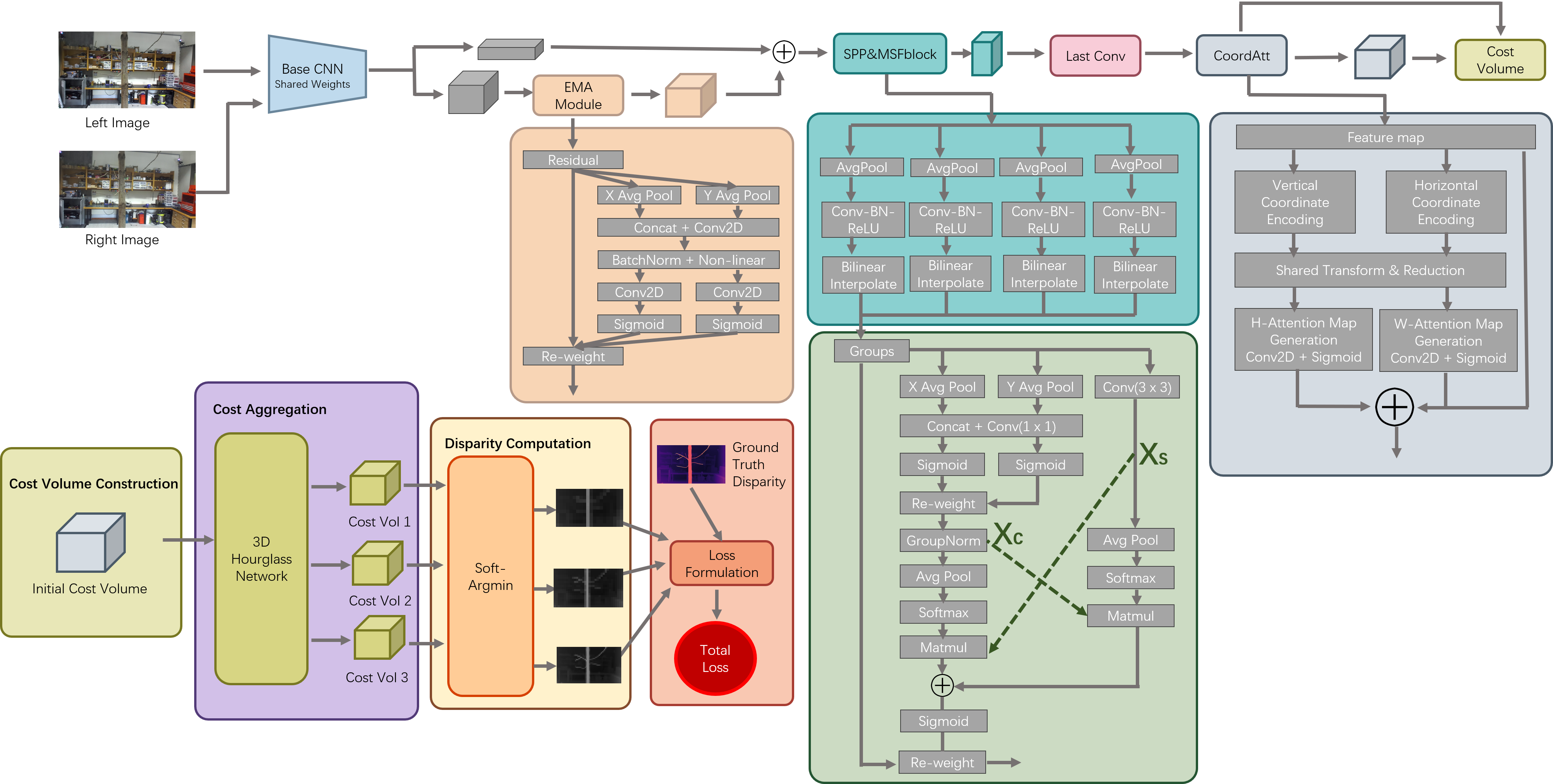}
        \caption{EMCStereo architecture. A weight-shared Base CNN extracts
        $1/4$-resolution features from both images. Along the main (top) path the
        deep $128$-channel feature is re-weighted by EMA, the four
        spatial-pyramid branches are adaptively fused by MSFblock, and the final
        $32$-channel feature is refined by CoordAtt before the concatenation cost
        volume is built. The volume is regularized by a stacked-hourglass 3D
        network and decoded by soft-argmin into three deeply supervised
        multi-scale predictions. The coloured insets detail the three modules.}
        \label{fig:arch}
    \end{figure*}

    EMCStereo follows the classic four-stage stereo pipeline (feature extraction,
    cost-volume construction, 3D aggregation, disparity regression) and enhances
    the feature stage with the three attention modules that give it its name
    (Fig.~\ref{fig:arch}). Given a rectified pair $(I_{L},
    I_{R})\in\mathbb{R}^{H\times W\times 3}$, a weight-shared backbone extracts
    $1/4$-resolution features $f_{L}, f_{R}\in\mathbb{R}^{32\times H/4\times
    W/4}$. A concatenation cost volume is formed across the disparity range,
    aggregated by a stacked-hourglass 3D network and regressed into a dense map
    $D\in\mathbb{R}^{H\times W}$, from which metric depth follows. The goal is to
    leave the PSMNet~\cite{chang2018psmnet} template intact and add attention
    only where it should help thin-structure matching most.

    \subsection{Attention-Enhanced Feature Backbone}

    The backbone is the PSMNet residual stem: three $3{\times}3$ convolutions
    down to $1/2$ resolution ($32$ channels) followed by four residual stages, of
    which \texttt{layer2} yields a $1/4$-resolution $64$-channel feature
    ($\mathrm{out}_{4}$) and \texttt{layer3}--\texttt{layer4} deepen it to $128$
    dilated channels ($\mathrm{out}_{8}$). We insert each module where its
    inductive bias fits.

    \textbf{EMA on the deep semantic feature.} Right after the dilated
    \texttt{layer4} we apply Efficient Multi-scale
    Attention~\cite{ouyang2023ema} to the $128$-channel feature. It splits the
    channels into $g=16$ groups and, per group, multiplies a coordinate branch
    (height and width pooling through a shared $1{\times}1$ convolution) with a
    parallel $3{\times}3$ branch, then normalizes the cross-spatial product by
    group normalization and softmax. On the deepest feature this strengthens
    long-range semantic context, so thin branches stay separable from background
    clutter. The module costs $672$ parameters.

    \textbf{MSFblock for adaptive pyramid fusion.} PSMNet pools
    $\mathrm{out}_{8}$ at four scales and concatenates the results. We instead
    fuse the four $32$-channel pyramid branches $\{b_{1},\dots,b_{4}\}$ with the
    multi-scale fusion block of SHISRCNet~\cite{chen2023shisrcnet}: each branch
    is globally pooled and projected to a channel descriptor, and the stacked
    descriptors pass through a sigmoid and a softmax to give normalized
    channel-wise weights $\{w_{i}\}$, so that $f_{\mathrm{msf}}=
    \mathrm{proj}(\sum_{i}w_{i}\odot b_{i})$ with $\odot$ channel-wise scaling.
    The network can then favour whichever pyramid scale matches the local branch
    width. The block adds $5{,}184$ parameters but collapses four branches into
    one, which is what makes the design a net saving
    (Section~\ref{subsec:cost}).

    \textbf{CoordAtt on the fused feature.} The fused feature is concatenated
    with $\mathrm{out}_{4}$ ($64$) and the EMA-enhanced $\mathrm{out}_{8}$
    ($128$) into a $224$-channel tensor, which \texttt{lastconv} compresses to
    $32$ channels. Coordinate Attention~\cite{hou2021coordatt} then pools
    separately along the height and width axes, encodes direction-aware maps
    $a_{h}$ and $a_{w}$, and re-scales as $f \leftarrow f\odot a_{h}\odot a_{w}$.
    Disparity varies smoothly along a scanline but sharply across a branch
    boundary, so this axis-factorized form suits the final matching feature
    ($856$ parameters).

    \subsection{Cost Volume, Aggregation and Loss}

    Shifting $f_{R}$ across the disparity range and stacking it with $f_{L}$
    gives a $64\times \tfrac{D_{\max}}{4}\times\tfrac{H}{4}\times\tfrac{W}{4}$
    concatenation volume with $D_{\max}=192$, regularized by the PSMNet
    stacked-hourglass aggregator (two entry $3{\times}3{\times}3$ residual
    blocks and three encoder--decoder hourglasses with skip connections). Three
    $1$-channel heads produce
    $c_{1},c_{2},c_{3}$, refined residually as
    $c_{2}\!\leftarrow\!c_{2}+c_{1}$ and $c_{3}\!\leftarrow\!c_{3}+c_{2}$, each
    upsampled to full resolution and turned into a disparity map by
    differentiable soft-argmin~\cite{kendall2017gcnet}, $\hat{D}=
    \sum_{d}d\cdot\mathrm{softmax}(c(d))$. We train with a deeply supervised
    smooth-$L_{1}$ loss over the valid pixels $\mathcal{M}=\{D_{\mathrm{gt}}\in(0,
    D_{\max})\}$,
    \begin{equation}
        \mathcal{L}= \sum_{k=1}^{3}\lambda_{k}\, \mathrm{smooth}_{L_1}\!\big(\hat
        {D}_{k}[\mathcal{M}], D_{\mathrm{gt}}[\mathcal{M}]\big), \label{eq:loss}
    \end{equation}
    with $(\lambda_{1},\lambda_{2},\lambda_{3})=(0.5,0.7,1.0)$. To stay stable
    under the steep disparity gradients of thin branches, predictions are clamped
    to $\pm 2 D_{\max}$ and the loss is capped. At inference only $\hat{D}_{3}$
    is used.

    \subsection{Parameter and Runtime Cost}
    \label{subsec:cost}

    The three modules are almost free. EMA adds $672$ parameters, CoordAtt $856$
    and MSFblock $5{,}184$. But because MSFblock replaces the four-branch pyramid
    concatenation ($4{\times}32$ channels) with a single $32$-channel fused
    tensor, the \texttt{lastconv} input shrinks from $320$ to $224$ channels and
    sheds $110{,}592$ parameters. The net effect is $-103{,}880$: $5{,}121{,}272$
    parameters against the identical backbone's $5{,}225{,}152$, or $2.0\%$
    smaller. On an NVIDIA RTX~3060 Laptop GPU (fp32, batch~1) a forward pass at
    $1248{\times}384$ takes $393$\,ms and peaks at $2.37$\,GiB, of which the
    three modules account for $3.31$\,ms per view, so $6.6$\,ms or $1.7\%$ per
    pair. The cost is dominated by the stacked-hourglass 3D convolutions, whose
    activations reach $10.27$\,GiB at full $1920{\times}1088$ resolution.
    Attention is not what stands between this design and real-time deployment;
    the cost volume is.

    \vspace{-0.85em}
    \section{The VirtualTree Dataset}

    \subsection{Rendering and Geometry}

    VirtualTree is rendered in Unreal Engine~5~\cite{ue5} with a virtual stereo
    rig that matches a ZED Mini camera~\cite{zedmini}: baseline $B=6.3$\,cm and
    focal length $f_{\mathrm{px}}=960$\,px (a $90^{\circ}$ horizontal field of
    view at $1920$\,px width). For each scene the engine exports left and right
    RGB images and a dense single-channel OpenEXR depth map. Because UE5 provides
    an exact geometry buffer, the ground truth is complete and boundary-accurate
    even for branches a few pixels wide, which is what nobody can label densely
    in the real world. Depth becomes disparity through the calibrated geometry,
    \begin{equation}
        D(x,y) = \frac{f_{\mathrm{px}}\,B}{Z(x,y)}, \label{eq:disp}
    \end{equation}
    with sky and invalid pixels (very large or non-finite depth) masked out so
    that they never contribute to training or evaluation.

    \subsection{Composition and Splits}

    The dataset holds $5{,}520$ stereo pairs rendered from $115$ distinct
    simulated trees, each captured from three viewpoint families (upward,
    downward and near-parallel), so that both near, wide branches and far, thin
    twigs appear. We divide the pairs $80/10/10$ into training, validation and
    test sets of $4{,}416$, $552$ and $552$ pairs and release the split. It is
    drawn at the frame level rather than the tree level, so different frames of
    one tree may fall on either side of it; the VirtualTree numbers below
    therefore measure generalization to unseen viewpoints of known trees, and we
    flag a tree-disjoint split as future work (Section~\ref{subsec:limits}).
    Training uses random $256\times512$ crops with mild, mostly symmetric stereo
    colour jitter; validation and testing use full-resolution images padded to a
    multiple of $32$.

    \vspace{-0.85em}
    \section{Experimental Setup}

    \subsection{Implementation Details}
    \label{subsec:impl}

    EMCStereo is implemented in PyTorch as a self-contained codebase. All models
    are trained with AdamW (learning rate $5\times10^{-4}$, weight decay
    $1\times10^{-4}$) under a cosine schedule with a $500$-iteration linear
    warm-up, automatic mixed precision, gradient clipping at $1.0$ and seed $0$
    (Table~\ref{tab:training}). The eight ablation variants of
    Section~\ref{subsec:ablation} are trained from scratch and share a fixed
    $100$-epoch budget, so they are compared at equal cost. KITTI~2012,
    KITTI~2015 and Middlebury are likewise trained from scratch on each
    benchmark's public training set (up to $300$ epochs, early-stopping patience
    $50$); only ETH3D, with $24$ training pairs, is instead a one-epoch
    fine-tune of the SceneFlow~\cite{mayer2016large} checkpoint.

    Two runs deviate, and we report them as trained rather than as planned. The
    headline VirtualTree model is \emph{not} trained from scratch: it warm-starts
    from an earlier VirtualTree checkpoint of the same architecture and then runs
    the full $300$-epoch schedule (best validation EPE at epoch $252$), so it has
    seen far more optimization than any ablation variant, a gap
    Section~\ref{subsec:budget} measures. The SceneFlow model, to fit the much
    larger corpus, uses a $288{\times}576$ crop, learning rate
    $4\times10^{-4}$ with a $200$-iteration warm-up, deep-supervision weights
    $(0.25,0.5,1.0)$, gradient accumulation to an effective batch of $24$ and
    weight EMA (decay $0.9995$); it ran as several resumed cosine segments, and
    the reported checkpoint is the EMA weights at epoch $108$. No test-time
    augmentation is used anywhere. One $100$-epoch VirtualTree run takes about
    $79$\,h on a single GPU of a shared cluster.

    \begin{table}[!tbp]
        \caption{Training configuration and measured cost for EMCStereo.
        Inference figures are fp32, batch~1, on an RTX~3060 Laptop GPU.}
        \label{tab:training}
        \centering
        \resizebox{\columnwidth}{!}{%
        \begin{tabular}{lc}
            \toprule \textbf{Setting} & \textbf{Value}                                  \\
            \midrule Optimizer        & AdamW ($\beta{=}0.9/0.999$, wd $10^{-4}$)       \\
            Learning rate             & $5\times10^{-4}$, cosine, 500-iter warm-up      \\
            Precision / clipping      & AMP; gradient clip $1.0$; seed $0$              \\
            Max disparity / crop      & $192$; $256\times512$; batch $4$                \\
            Colour augmentation       & Jitter $0.4/0.4/0.4/0.16$, asymmetric $p{=}0.2$ \\
            Epochs (VirtualTree)      & 300, warm-started from a VirtualTree ckpt        \\
            Epochs (ablation)         & fixed 100 from scratch (patience 50)            \\
            KITTI / Middlebury        & from scratch, up to 300 (patience 50)           \\
            ETH3D                     & 1-epoch fine-tune of SceneFlow ckpt             \\
            SceneFlow                 & $288{\times}576$, eff.\ batch 24, EMA, ep.\ 108    \\
            Parameters                & $5{,}121{,}272$ ($5{,}225{,}152$ w/o attention)  \\
            Latency / peak mem.\ @\,$1248{\times}384$ & $393$\,ms; $2.37$\,GiB          \\
            Peak mem.\ @\,$1920{\times}1088$          & $10.27$\,GiB                    \\
            \bottomrule
        \end{tabular}%
        }
    \end{table}

    \subsection{Evaluation Metrics}

    For disparity we report EPE (the mean absolute disparity error), RMSE, the
    KITTI-style D1-all (error $>3$\,px \emph{and} $>5\%$) and the Bad-$\tau$
    rates for $\tau\in\{1,3\}$\,px. For depth, obtained by inverting
    Eq.~\eqref{eq:disp}, we report absolute relative error (AbsRel) and the
    threshold accuracy $\delta_{1}$ ($\max(\hat{Z}/Z, Z/\hat{Z}) < 1.25$). Only
    valid ground-truth pixels count, that is, finite and in $(0, D_{\max})$, and
    padded images are un-padded before scoring, so no padded pixel enters any
    metric. One protocol point matters throughout: VirtualTree is the only
    benchmark whose evaluation split is disjoint from checkpoint selection, so it
    alone gives an unbiased number. The ablation and the four standard benchmarks
    score the best-validation checkpoint on that same validation split, and are
    selection-biased upper bounds.

    \vspace{-0.85em}
    \section{Results}

    \subsection{Main Results on VirtualTree}

    In Table~\ref{tab:virtualtree} the validation and the untouched test split
    agree to within $0.007$\,px EPE and $0.16$\,pp D1-all, so checkpoint
    selection buys nothing and the held-out numbers can be read at face value.
    Thin-branch geometry is unforgiving: one misassigned pixel can jump from a
    foreground twig to the far background. EMCStereo still reaches $1.31$\,px EPE
    at $5.96\%$ D1-all on the test split, so about one pixel in seventeen
    violates the KITTI $3$-px$/5\%$ criterion. The Bad-$1.0$ rate of $15.16\%$
    reflects the near-boundary pixels that dominate branch scenes, but half of
    those are already within $3$\,px (Bad-$3.0$ is $7.31\%$): gross mismatches
    are rare, and the residual error sits in a narrow band around the
    silhouettes. In the depth domain the same prediction gives AbsRel $0.056$ and
    $\delta_{1}=96.1\%$.

    \begin{table}[!tbp]
        \caption{EMCStereo on the two VirtualTree splits ($552$ pairs each). The
        validation split selects the checkpoint; the test split takes part in
        neither training nor selection.}
        \label{tab:virtualtree}
        \centering
        \resizebox{\columnwidth}{!}{%
        \begin{tabular}{lcccccc}
            \toprule \textbf{Split} & \textbf{EPE}\,$\downarrow$ & \textbf{D1-all}\,$\downarrow$ & \textbf{Bad 1.0}\,$\downarrow$ & \textbf{Bad 3.0}\,$\downarrow$ & \textbf{RMSE}\,$\downarrow$ & \textbf{$\delta_{1}$}\,$\uparrow$ \\
                                    & (px)                       & (\%)                          & (\%)                           & (\%)                           & (px)                        & (\%)                              \\
            \midrule Validation     & 1.315                      & 5.80                          & 15.33                          & 7.33                           & 4.83                        & 96.20                             \\
            Test (held out)         & 1.308                      & 5.96                          & 15.16                          & 7.31                           & 4.71                        & 96.11                             \\
            \bottomrule
        \end{tabular}%
        }
    \end{table}

    \subsection{The Training Budget Dominates}
    \label{subsec:budget}

    The headline model and the ablation variants differ in budget as well as in
    architecture, so the two must be separated before the ablation is read. At
    the matched $100$-epoch budget the full model reaches $1.752$\,px on the
    VirtualTree validation split, against $1.749$ and $1.740$\,px for the
    no-attention baseline in two independent runs: indistinguishable. Given a
    $300$-epoch schedule instead, the same from-scratch model keeps improving, to
    $1.726$\,px at epoch $116$ and $1.674$\,px at epoch $138$, where the run
    stood at submission. Two extra epochs alone, a warm restart at
    $2\times10^{-4}$ from the $1.726$\,px checkpoint and therefore free of any
    run-to-run confound, take it to $1.644$\,px ($1.606$\,px on the test split).
    That one step is worth $0.082$\,px, more than the $0.056$\,px between the
    best and the worst of the eight ablation variants. The headline model, which
    warm-starts and completes the full schedule, ends at $1.315$\,px. Schedule
    length, not architecture, is the dominant variable here, which is the main
    reason to read the grid conservatively.

    \subsection{Ablation Study}
    \label{subsec:ablation}

    To isolate each module we train a separate model for every combination of
    EMA, MSFblock and CoordAtt on VirtualTree. All eight share one protocol, one
    split, one seed and one $100$-epoch budget, so they are mutually comparable,
    but only with each other, since the split that scores them also selected
    their checkpoints. The final row of Table~\ref{tab:ablation} steps outside
    that grid to show the deployed model at its own budget.

    \textbf{A measured noise floor.} Ablation margins in stereo matching are
    routinely quoted at the third decimal of EPE, yet a cost-volume network
    trained under mixed precision with non-deterministic cuDNN kernels is not
    reproducible to that precision. We therefore re-ran the no-attention baseline
    end to end, same configuration and seed but different hardware and an
    independently written training driver, and take the disagreement as the
    resolution of the study. The two runs land at $1.749$ and $1.740$\,px: a
    spread of $0.009$\,px EPE, $0.03$\,pp D1-all, $0.11$\,pp Bad-$1.0$ and
    $0.059$\,px RMSE. Margins below roughly $0.01$\,px EPE are therefore not
    results. The floor comes from training, not from measurement: re-evaluating
    the \emph{identical} checkpoint on different hardware reproduces $1.740427$
    against $1.740423$\,px, a gap of $4\times10^{-6}$\,px, with the other three
    metrics identical to six decimals.

    \textbf{What survives the floor.} At $100$ epochs the eight variants form two
    groups rather than a ranking. The two baseline runs and all four
    configurations containing EMA occupy $1.736$--$1.752$\,px, a span of
    $0.016$\,px, so neither the full model ($1.752$\,px) nor the lowest single
    number, EMA+CoordAtt ($1.736$\,px), is separable from the baseline or from
    each other. The three configurations that use MSFblock or CoordAtt
    \emph{without} EMA form a second group at $1.777$--$1.792$\,px,
    $0.03$--$0.05$\,px behind the baseline. That gap is three to five times the
    floor, holds across all four metrics and is mirrored by RMSE
    ($5.288$--$5.365$ against $5.401$--$5.435$\,px). It is the one effect this
    ablation establishes: MSFblock and CoordAtt cost accuracy on their own, and
    EMA is what makes them harmless, plausibly by leaving the deep feature
    discriminative enough that the later re-weighting acts on signal. At a
    matched budget the attention stack therefore does not beat a plain
    PSMNet-style backbone on accuracy; what it buys is the $2.0\%$ smaller model
    of Section~\ref{subsec:cost}. We keep all three modules because they cost
    nothing and because the deployed model runs the longer schedule, at which the
    variants have not been compared.

    \begin{table}[!tbp]
        \caption{Ablation of the three attention modules on the VirtualTree
        validation split ($552$ pairs); lower is better. The eight upper rows
        share a fixed $100$-epoch budget from scratch and are mutually
        comparable; an independent repeat of the top row gave $1.740$\,px, so
        differences below $0.009$\,px EPE ($0.03$\,pp D1-all, $0.11$\,pp
        Bad-$1.0$, $0.059$\,px RMSE) sit inside the noise floor. The last row is
        the deployed model of Table~\ref{tab:virtualtree}, whose margin is a
        budget effect (Section~\ref{subsec:budget}), not an architectural one.}
        \label{tab:ablation}
        \centering
        \resizebox{\columnwidth}{!}{%
        \begin{tabular}{ccccccccc}
            \toprule \textbf{EMA} & \textbf{MSF} & \textbf{Coord} & \textbf{Params}  & \textbf{EPE}\,$\downarrow$ & \textbf{D1-all}\,$\downarrow$ & \textbf{Bad 1.0}\,$\downarrow$ & \textbf{RMSE}\,$\downarrow$ \\
                                  &              &                & (M)                          & (px)                       & (\%)                          & (\%)                           & (px)                        \\
            \midrule              &              &                & 5.225                     & 1.749                      & 8.01                          & 19.73                          & 5.346                       \\
            \checkmark            &              &                & 5.226                     & 1.740                      & 8.07                          & 19.99                          & 5.300                       \\
                                  & \checkmark   &                & 5.120                     & 1.777                      & 8.10                          & 19.87                          & 5.401                       \\
                                  &              & \checkmark     & 5.226                     & 1.785                      & 8.15                          & 19.97                          & 5.407                       \\
            \checkmark            & \checkmark   &                & 5.120                     & 1.738                      & 8.00                          & 19.79                          & 5.318                       \\
            \checkmark            &              & \checkmark     & 5.227                     & 1.736                      & 8.07                          & 19.73                          & 5.288                       \\
                                  & \checkmark   & \checkmark     & 5.121                     & 1.792                      & 8.20                          & 20.05                          & 5.435                       \\
            \checkmark            & \checkmark   & \checkmark     & 5.121                     & 1.752                      & 8.08                          & 19.90                          & 5.365                       \\
            \midrule \multicolumn{3}{l}{\textit{Deployed model (300 ep.)}}      & 5.121                     & \textbf{1.315}             & \textbf{5.80}                 & \textbf{15.33}                 & \textbf{4.826}              \\
            \bottomrule
        \end{tabular}%
        }
    \end{table}

    \begin{figure*}[!t]
        \centering
        \includegraphics[width=\textwidth]{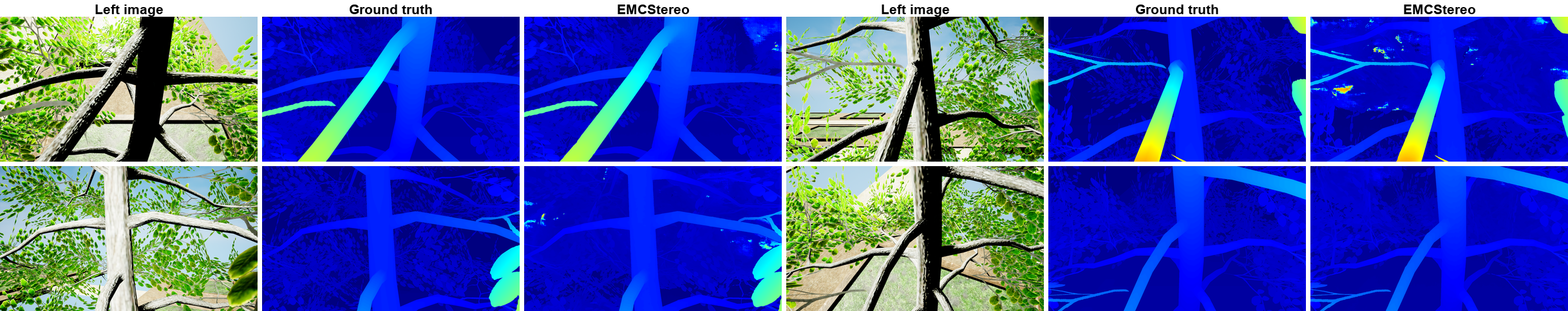}
        \caption{Four VirtualTree validation scenes, two per row; each scene
        shows the left RGB frame, the ground-truth disparity from the UE5
        geometry buffer and the EMCStereo prediction, with the disparity panels
        sharing the JET map over $[0, D_{\max}]$. The top row holds two views
        with a near branch spanning much of the search range, the bottom row two
        canopy views dominated by thin twigs. Twigs stay connected and
        silhouettes stay sharp; the residual speckle sits in the sky and
        background foliage, which the metrics exclude.}
        \label{fig:qual_synth}
    \end{figure*}

    \begin{figure}[!tbp]
        \centering
        \setlength{\tabcolsep}{0pt}
        \begin{tabular}{@{}*{3}{>{\centering\arraybackslash}m{0.33\columnwidth}}@{}}
            \includegraphics[width=0.31\columnwidth]{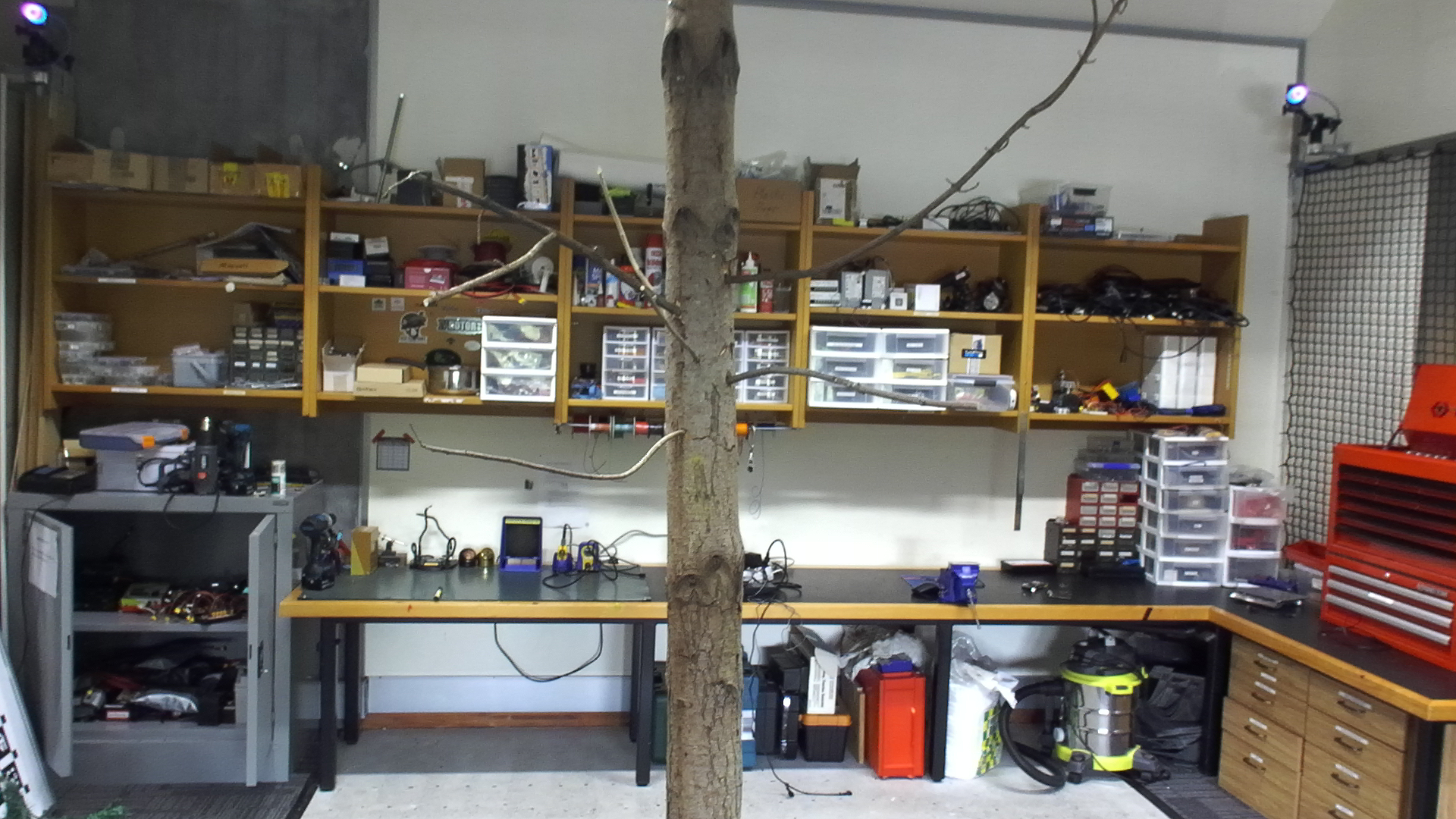} & \includegraphics[width=0.31\columnwidth]{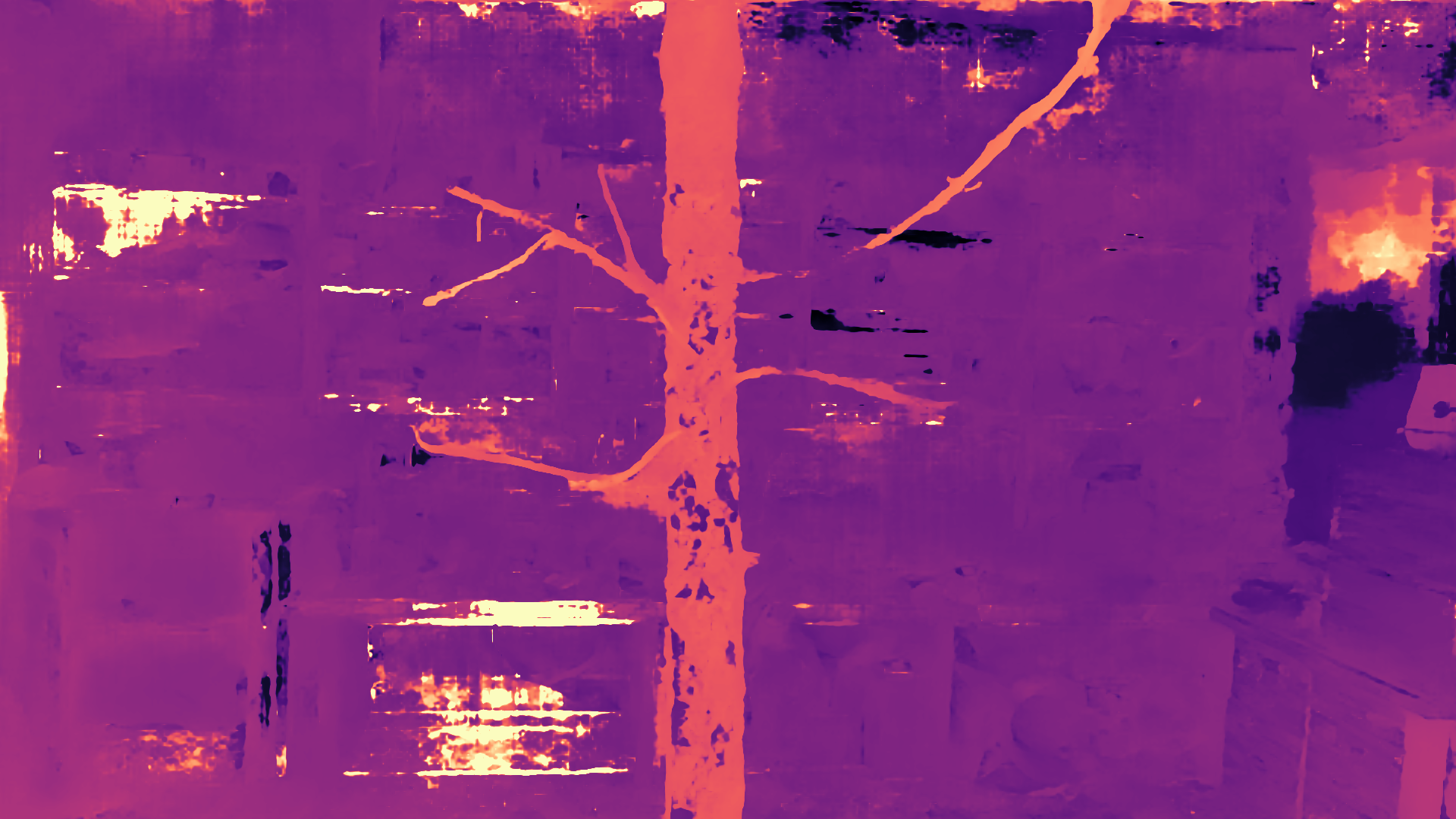} & \includegraphics[width=0.31\columnwidth]{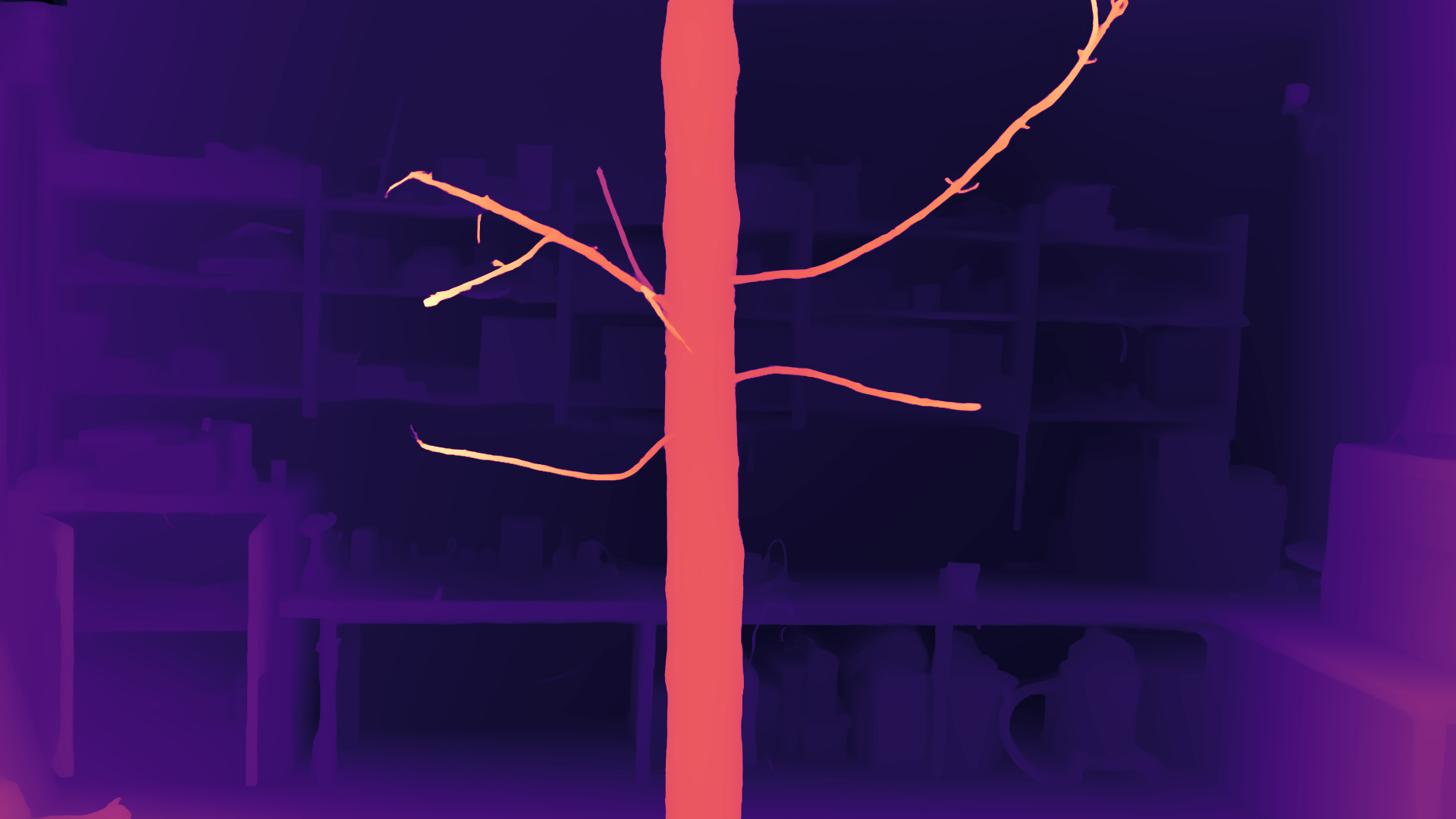} \\
            \footnotesize Left input                                      & \footnotesize EMCStereo (ours)                               & \footnotesize DEFOM-Stereo~\cite{jiang2025defom}                \\
        \end{tabular}
        \caption{Real-world stereo pair (physical ZED Mini). Trained only on
        synthetic data, EMCStereo recovers the trunk, the branches and their
        depth order; the cluttered indoor background is far outside its training
        domain. Our map is colour-coded over $0$--$50$\,px and the reference uses
        its own normalization, so only structure is comparable.}
        \label{fig:qual_real}
    \end{figure}

    \begin{figure}[!tbp]
        \centering
        \includegraphics[width=\columnwidth]{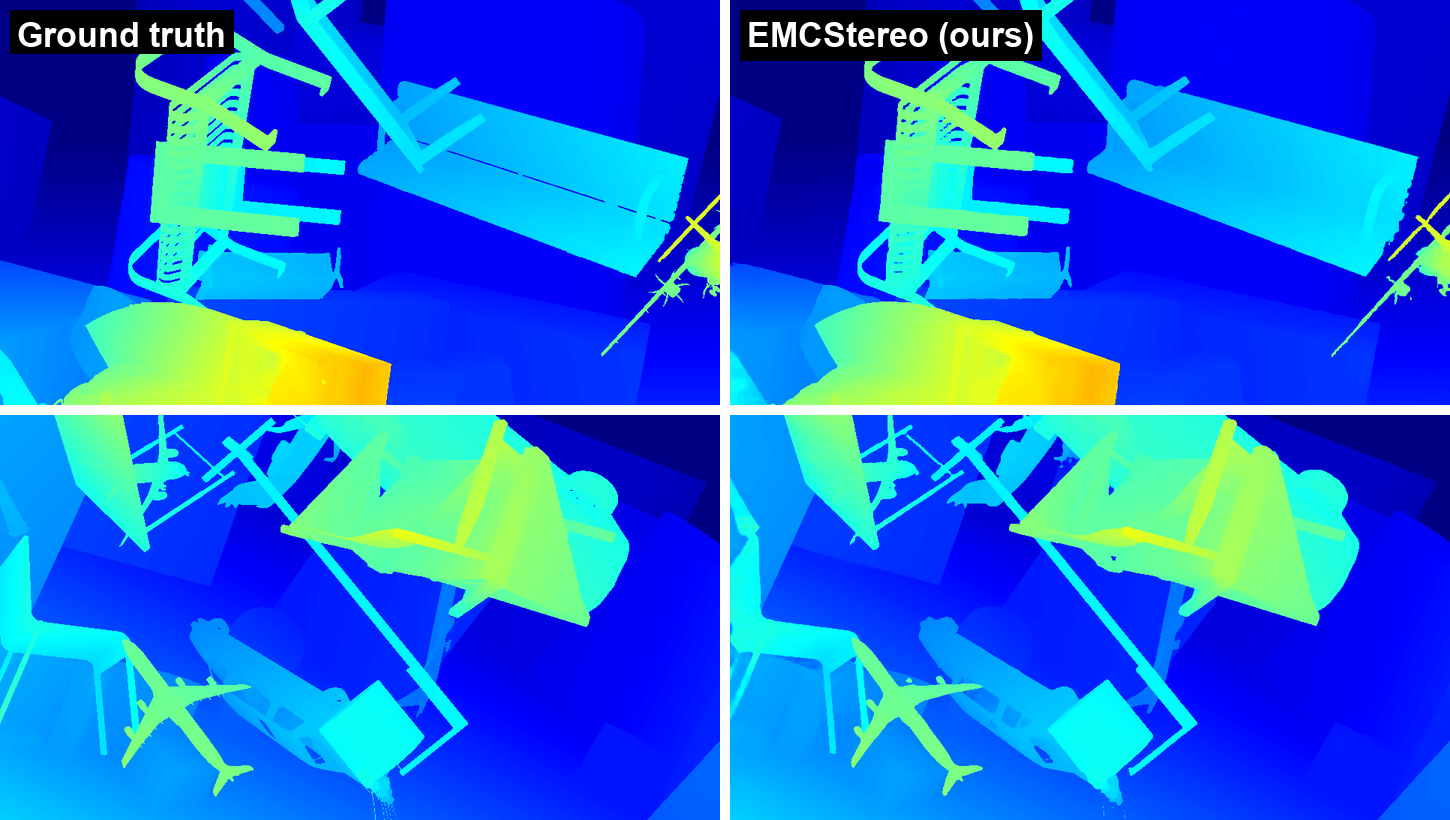}
        \caption{Two scenes from the held-out SceneFlow split: ground truth
        (left) against EMCStereo (right), JET-coloured over $[0, D_{\max}]$. Wire
        racks, chair legs and aircraft edges survive at close to their
        ground-truth width, and most of the residual error is in the thin
        occlusion bands the ground truth marks invalid.}
        \label{fig:qual_sf}
    \end{figure}

    \subsection{Cross-Dataset Evaluation}
    \label{subsec:cross}

    To check that EMCStereo is not tuned to synthetic branches, we also train it
    on SceneFlow, KITTI~2012, KITTI~2015, ETH3D and Middlebury
    (Table~\ref{tab:cross}). All are trained from scratch on their public
    training sets except ETH3D, which has only $24$ training pairs and is a
    one-epoch fine-tune of the SceneFlow checkpoint. Only the public training
    sets were available, so each is split $90/10$ (seed $0$) and the held-out
    part serves as both early-stopping signal and evaluation set. On the four
    real benchmarks that split is tiny ($19$, $20$, $3$ and $5$ pairs) and
    selection-biased, so those rows show that the architecture trains and
    generalizes on real imagery; they are not leaderboard measurements. The
    SceneFlow row is the exception: its held-out split holds $3{,}545$ pairs, so
    it is a stable estimate, even though the checkpoint was still chosen on it.

    Within those limits EMCStereo transfers well. On SceneFlow it reaches
    $1.00$\,px EPE at $3.59\%$ D1-all and $\delta_{1}=97.67\%$, with only
    $8.91\%$ of pixels off by more than $1$\,px. It reaches sub-pixel EPE on the
    two driving benchmarks ($0.80$\,px on KITTI~2012 and $0.73$\,px on
    KITTI~2015, with D1-all of $3.25\%$ and $2.20\%$) and $0.62$\,px on the
    high-resolution, low-texture ETH3D. Middlebury, whose disparities press
    hardest against the $D_{\max}=192$ search range, is the hardest of the five
    at $3.19$\,px EPE and $12.65\%$ D1-all. Depth accuracy holds up throughout,
    with $\delta_{1}$ from $92.6\%$ on ETH3D to $98.7\%$ on KITTI~2012 and AbsRel
    at or below $0.098$, which is what matters downstream, where a controller
    consumes metric distance rather than raw disparity. Surviving five domains
    that differ in resolution, disparity range and scene type suggests the
    backbone learns general matching features rather than dataset-specific
    shortcuts. It does not show that the design improves on the PSMNet family;
    only the ablation could carry that claim, and it does not.

    \begin{table}[!tbp]
        \caption{Cross-dataset evaluation of EMCStereo trained on each
        benchmark's public training set (all from scratch; ETH3D as a one-epoch
        fine-tune of the SceneFlow checkpoint). ``Pairs'' is the size of the
        $10\%$ held-out split, which also supplies the early-stopping signal, so
        the metrics are selection-biased upper bounds rather than leaderboard
        results.}
        \label{tab:cross}
        \centering
        \resizebox{\columnwidth}{!}{%
        \begin{tabular}{lcccccc}
            \toprule \textbf{Dataset} & \textbf{Pairs} & \textbf{EPE}\,$\downarrow$ & \textbf{D1-all}\,$\downarrow$ & \textbf{RMSE}\,$\downarrow$ & \textbf{AbsRel}\,$\downarrow$ & \textbf{$\delta_{1}$}\,$\uparrow$ \\
                                      &                & (px)                       & (\%)                          & (px)                        &                               & (\%)                              \\
            \midrule SceneFlow        & 3{,}545        & 1.00                       & 3.59                          & 5.13                        & 0.097                         & 97.67                             \\
            KITTI~2012                & 19             & 0.80                       & 3.25                          & 2.79                        & 0.022                         & 98.74                             \\
            KITTI~2015                & 20             & 0.73                       & 2.20                          & 1.94                        & 0.027                         & 98.63                             \\
            ETH3D                     & 3              & 0.62                       & 3.86                          & 1.50                        & 0.078                         & 92.60                             \\
            Middlebury                & 5              & 3.19                       & 12.65                         & 9.43                        & 0.050                         & 96.17                             \\
            \bottomrule
        \end{tabular}%
        }
    \end{table}

    \begin{table}[!tbp]
        \caption{Comparison with published stereo networks, ordered by KITTI~2015
        D1-all. Accuracy is quoted from the cited work; parameter counts come
        from the cited work where it reports one, otherwise from a published
        comparison (DispNetC~\cite{tulyakov2018pds}, GwcNet and
        ACVNet~\cite{xu2022acvnet}, DeepPruner-Best~\cite{jeon2022msffnet});
        ``---'' marks a number none of these sources gives. Our row is
        \emph{not} a leaderboard entry and is printed at the precision we
        measured: $^{\dagger}$~$10\%$ held-out SceneFlow split ($3{,}545$ pairs);
        $^{\ddagger}$~$20$-pair held-out KITTI~2015 split. Only the parameter
        column is comparable across the two blocks.}
        \label{tab:sota}
        \centering
        \resizebox{\columnwidth}{!}{%
        \begin{tabular}{lccc}
            \toprule \textbf{Method}                    & \textbf{SceneFlow}     & \textbf{KITTI 2015}       & \textbf{Params} \\
                                                        & EPE (px)\,$\downarrow$ & D1-all (\%)\,$\downarrow$ & (M)             \\
            \midrule \multicolumn{4}{l}{\textit{Published, official test splits}}                                            \\
            DispNetC~\cite{mayer2016large}              & 1.68                   & 4.34                      & 42              \\
            MC-CNN-acrt~\cite{zbontar2016mccnn}         & ---                    & 3.88                      & ---             \\
            GC-Net~\cite{kendall2017gcnet}              & 2.51                   & 2.87                      & 2.85            \\
            CRL~\cite{pang2017crl}                      & 1.32                   & 2.67                      & 78.77           \\
            AANet$/$AANet+~\cite{xu2020aanet}           & 0.87$/$0.72            & 2.55$/$2.03               & 3.9$/$8.4       \\
            iResNet-i2~\cite{liang2018iresnet}          & 1.40                   & 2.44                      & 43.34           \\
            PSMNet~\cite{chang2018psmnet}               & 1.09                   & 2.32                      & 5.22            \\
            SegStereo~\cite{yang2018segstereo}          & 1.45                   & 2.25                      & ---             \\
            DeepPruner-Best~\cite{duggal2019deeppruner} & 0.86                   & 2.15                      & 7.39            \\
            GwcNet~\cite{guo2019gwcnet}                 & 0.76                   & 2.11                      & 6.91            \\
            ACVNet~\cite{xu2022acvnet}                  & 0.48                   & 1.65                      & 6.22            \\
            \midrule \multicolumn{4}{l}{\textit{This work, own splits (not leaderboard results)}}                            \\
            EMCStereo (ours)                            & 1.0050$^{\dagger}$     & 2.1955$^{\ddagger}$       & 5.1213          \\
            \bottomrule
        \end{tabular}%
        }
    \end{table}

    \subsection{Comparison with Published Methods}
    \label{subsec:compare}

    Table~\ref{tab:sota} places EMCStereo alongside eleven published networks;
    PSMNet is the closest reference, since our backbone derives from it. The
    accuracy columns are indicative only, because the published rows are official
    test-set results and ours are not: our SceneFlow figure evaluates the best
    EMA checkpoint (epoch $108$) on a $10\%$ held-out split of the suite
    ($3{,}545$ of $35{,}454$ pairs) under the recipe of
    Section~\ref{subsec:impl}, and our KITTI~2015 D1-all is the $20$-pair
    validation figure of Table~\ref{tab:cross}. Both are selection-biased upper
    bounds, so EMCStereo landing below PSMNet's $1.09$\,px on SceneFlow, and
    above seven of the eleven published rows on KITTI~2015 D1-all, is not a claim
    of superiority. The parameter column is protocol-independent, and there the
    comparison does carry: at $5.12$\,M, EMCStereo is an order of magnitude
    smaller than the refinement-cascade networks in the same accuracy band
    ($78.77$\,M for CRL, $43.34$\,M for iResNet-i2), smaller than every
    cost-volume network in the table except GC-Net, and $2.0\%$ below the same
    backbone without attention. A controlled re-benchmarking of every method
    under one protocol is left to future work.

    \subsection{Qualitative Results}
    \label{subsec:qual}

    In Fig.~\ref{fig:qual_synth} thin branches stay connected and
    branch--background boundaries stay sharp, which is where over-smoothing shows
    up first, and the near branches of the top row are recovered across most of
    the search range; that is the visual counterpart of the $7.31\%$ Bad-$3.0$
    rate in Table~\ref{tab:virtualtree}. Fig.~\ref{fig:qual_sf} shows the same
    behaviour away from vegetation: on the held-out SceneFlow split the wire rack
    and the chair legs keep their width and their separation from the background,
    and the visible errors are concentrated in the occlusion bands that the
    ground truth itself marks invalid. On the real pair of
    Fig.~\ref{fig:qual_real}, EMCStereo, trained only on synthetic foliage, still
    separates the trunk and every major branch at the correct depth order, which
    is what the pruning application needs. The out-of-domain indoor background is
    reconstructed less cleanly than in the DEFOM-Stereo~\cite{jiang2025defom}
    reference, which carries a monocular depth prior trained on real imagery;
    with no ground truth there, this is a reference, not a measurement.

    \vspace{-0.85em}
    \section{Limitations and Future Work}
    \label{subsec:limits}

    \textit{Budget} dominates everything else we measured
    (Section~\ref{subsec:budget}): the headline model warm-starts and trains for
    $300$ epochs while the ablation grid runs $100$ from scratch, so a
    no-attention baseline under the headline schedule is the most valuable
    missing experiment, and several seeds would turn our single repeat into a
    confidence interval. \textit{Selection bias}: the ablation and the four
    cross-domain benchmarks score the best-validation checkpoint on the
    validation split itself, over as few as $3$--$20$ pairs, so those numbers are
    upper bounds. \textit{Split construction}: the VirtualTree split is per
    frame, not per tree, so it measures generalization to unseen viewpoints
    rather than unseen trees; a tree-disjoint re-split will be released alongside
    it. \textit{Domain gap}: Fig.~\ref{fig:qual_real} shows the consequence of
    not modelling lighting, sensor noise and wind, so real-domain or
    self-supervised adaptation is a natural extension. \textit{Compute}: the 3D
    aggregator dominates, at $393$\,ms per $1248{\times}384$ pair against
    $6.6$\,ms for all three attention modules, so lighter aggregation is the
    route to real-time deployment and to coupling EMCStereo with branch detection
    as an end-to-end distance estimator for autonomous pruning.

    \vspace{-0.85em}
    \section{Conclusion}

    We presented EMCStereo, which adds EMA, MSFblock and CoordAtt to a compact
    PSMNet-style cost-volume backbone, and VirtualTree, a UE5 stereo dataset of
    $5{,}520$ pairs with dense, exact ground-truth disparity for tree branches.
    EMCStereo reaches $1.31$\,px EPE at $5.96\%$ D1-all on the held-out test
    split, $1.00$\,px on a held-out SceneFlow split, and $\delta_{1}$ between
    $92.6\%$ and $98.7\%$ on four standard real-image benchmarks. We also report
    a negative result: an eight-way ablation with an independent repeat of the
    baseline puts run-to-run variability at $0.009$\,px EPE, and at a matched
    budget the attention stack sits inside that floor, accuracy-neutral but
    $2.0\%$ smaller. Stating that floor lets later work be measured against it.
    VirtualTree, the code and every run's artefacts are released with the paper.

    \FloatBarrier

    \bibliographystyle{IEEEtran}

\begin{thebibliography}{99}
        \bibitem{lin2024branch} Y. Lin, B. Xue, M. Zhang, S. Schofield, and R.
            Green, ``Deep learning-based depth map generation and YOLO-integrated
            distance estimation for radiata pine branch detection using drone stereo
            vision,'' in \textit{Proc. Int. Conf. Image Vis. Comput. New Zealand
            (IVCNZ)}, Christchurch, 2024, pp. 1--6.

        \bibitem{lin2025segmentation} Y. Lin, B. Xue, M. Zhang, S. Schofield, and
            R. Green, ``Performance evaluation of deep learning for tree branch segmentation
            in autonomous forestry systems,'' in \textit{Proc. Int. Conf. Image Vis.
            Comput. New Zealand (IVCNZ)}, Wellington, 2025, pp. 1--6.

        \bibitem{lin2025yolosgbm} Y. Lin, B. Xue, M. Zhang, S. Schofield, and R.
            Green, ``YOLO and SGBM integration for autonomous tree branch
            detection and depth estimation in radiata pine pruning
            applications,'' in \textit{Proc. Int. Conf. Image Vis. Comput. New
            Zealand (IVCNZ)}, Wellington, 2025, pp. 1--6.

        \bibitem{lin2025genetic} Y. Lin, B. Xue, M. Zhang, S. Schofield, and R.
            Green, ``Genetic algorithms for parameter optimization for disparity
            map generation of radiata pine branch images,'' in \textit{Proc. Int.
            Conf. Image Vis. Comput. New Zealand (IVCNZ)}, Wellington, 2025, pp.
            1--6.

        \bibitem{kendall2017gcnet} A. Kendall, H. Martirosyan, S. Dasgupta, P.
            Henry, R. Kennedy, A. Bachrach, and A. Bry, ``End-to-end learning of
            geometry and context for deep stereo regression,'' in \textit{Proc. IEEE
            Int. Conf. Comput. Vis. (ICCV)}, 2017, pp. 66--75.

        \bibitem{chang2018psmnet} J.-R. Chang and Y.-S. Chen, ``Pyramid stereo
            matching network,'' in \textit{Proc. IEEE Conf. Comput. Vis. Pattern
            Recognit. (CVPR)}, 2018, pp. 5410--5418.

        \bibitem{guo2019gwcnet} X. Guo, K. Yang, W. Yang, X. Wang, and H. Li, ``Group-wise
            correlation stereo network,'' in \textit{Proc. IEEE Conf. Comput. Vis.
            Pattern Recognit. (CVPR)}, 2019, pp. 3273--3282.

        \bibitem{zhang2019ganet} F. Zhang, V. Prisacariu, R. Yang, and P. H. S.
            Torr, ``GA-Net: Guided aggregation net for end-to-end stereo
            matching,'' in \textit{Proc. IEEE Conf. Comput. Vis. Pattern
            Recognit. (CVPR)}, 2019, pp. 185--194.

        \bibitem{xu2020aanet} H. Xu and J. Zhang, ``AANet: Adaptive aggregation network
            for efficient stereo matching,'' in \textit{Proc. IEEE Conf. Comput.
            Vis. Pattern Recognit. (CVPR)}, 2020, pp. 1959--1968.

        \bibitem{duggal2019deeppruner} S. Duggal, S. Wang, W.-C. Ma, R. Hu, and
            R. Urtasun, ``DeepPruner: Learning efficient stereo matching via differentiable
            PatchMatch,'' in \textit{Proc. IEEE Int. Conf. Comput. Vis. (ICCV)},
            2019, pp. 4384--4393.

        \bibitem{zbontar2016mccnn} J. \v{Z}bontar and Y. LeCun, ``Stereo matching
            by training a convolutional neural network to compare image
            patches,'' \textit{J. Mach. Learn. Res.}, vol. 17, no. 65, pp. 1--32,
            2016.

        \bibitem{pang2017crl} J. Pang, W. Sun, J. S. J. Ren, C. Yang, and Q.
            Yan, ``Cascade residual learning: A two-stage convolutional neural
            network for stereo matching,'' in \textit{Proc. IEEE Int. Conf.
            Comput. Vis. Workshops (ICCVW)}, 2017, pp. 878--886.

        \bibitem{liang2018iresnet} Z. Liang, Y. Feng, Y. Guo, H. Liu, W. Chen,
            L. Qiao, L. Zhou, and J. Zhang, ``Learning for disparity estimation
            through feature constancy,'' in \textit{Proc. IEEE Conf. Comput. Vis.
            Pattern Recognit. (CVPR)}, 2018, pp. 2811--2820.

        \bibitem{yang2018segstereo} G. Yang, H. Zhao, J. Shi, Z. Deng, and J.
            Jia, ``SegStereo: Exploiting semantic information for disparity
            estimation,'' in \textit{Proc. Eur. Conf. Comput. Vis. (ECCV)}, 2018,
            pp. 636--651.

        \bibitem{xu2022acvnet} G. Xu, J. Cheng, P. Guo, and X. Yang, ``Attention
            concatenation volume for accurate and efficient stereo matching,'' in
            \textit{Proc. IEEE/CVF Conf. Comput. Vis. Pattern Recognit. (CVPR)},
            2022, pp. 12971--12980.

        \bibitem{lipson2021raft} L. Lipson, Z. Teed, and J. Deng, ``RAFT-Stereo:
            Multilevel recurrent field transforms for stereo matching,'' in \textit{Proc.
            Int. Conf. 3D Vis. (3DV)}, 2021, pp. 218--227.

        \bibitem{jiang2025defom} H. Jiang, Z. Lou, L. Ding, R. Xu, M. Tan, W.
            Jiang, and R. Huang, ``DEFOM-Stereo: Depth foundation model based
            stereo matching,'' in \textit{Proc. IEEE/CVF Conf. Comput. Vis.
            Pattern Recognit. (CVPR)}, 2025, pp. 21857--21867.

        \bibitem{tulyakov2018pds} S. Tulyakov, A. Ivanov, and F. Fleuret,
            ``Practical deep stereo (PDS): Toward applications-friendly deep
            stereo matching,'' in \textit{Adv. Neural Inf. Process. Syst.
            (NeurIPS)}, 2018, pp. 5875--5885.

        \bibitem{jeon2022msffnet} S. Jeon and Y. S. Heo, ``Efficient multi-scale
            stereo-matching network using adaptive cost volume filtering,''
            \textit{Sensors}, vol. 22, no. 15, art. 5500, 2022.

        \bibitem{hu2018senet} J. Hu, L. Shen, and G. Sun, ``Squeeze-and-excitation
            networks,'' in \textit{Proc. IEEE Conf. Comput. Vis. Pattern
            Recognit. (CVPR)}, 2018, pp. 7132--7141.

        \bibitem{woo2018cbam} S. Woo, J. Park, J.-Y. Lee, and I. S. Kweon, ``CBAM:
            Convolutional block attention module,'' in \textit{Proc. Eur. Conf. Comput.
            Vis. (ECCV)}, 2018, pp. 3--19.

        \bibitem{hou2021coordatt} Q. Hou, D. Zhou, and J. Feng, ``Coordinate attention
            for efficient mobile network design,'' in \textit{Proc. IEEE Conf.
            Comput. Vis. Pattern Recognit. (CVPR)}, 2021, pp. 13713--13722.

        \bibitem{ouyang2023ema} D. Ouyang, S. He, G. Zhang, M. Luo, H. Guo, J.
            Zhan, and Z. Huang, ``Efficient multi-scale attention module with cross-spatial
            learning,'' in \textit{Proc. IEEE Int. Conf. Acoust., Speech, Signal
            Process. (ICASSP)}, 2023, pp. 1--5.

        \bibitem{chen2023shisrcnet} L. Xie, C. Li, Z. Wang, X. Zhang, B. Chen,
            Q. Shen, and Z. Wu, ``SHISRCNet: Super-resolution and classification
            network for low-resolution breast cancer histopathology image,'' in
            \textit{Proc. Int. Conf. Med. Image Comput. Comput.-Assist. Interv.
            (MICCAI)}, 2023, pp. 23--32.

        \bibitem{mayer2016large} N. Mayer, E. Ilg, P. Hausser, P. Fischer, D.
            Cremers, A. Dosovitskiy, and T. Brox, ``A large dataset to train convolutional
            networks for disparity, optical flow, and scene flow estimation,''
            in \textit{Proc. IEEE Conf. Comput. Vis. Pattern Recognit. (CVPR)},
            2016, pp. 4040--4048.

        \bibitem{geiger2012kitti} A. Geiger, P. Lenz, and R. Urtasun, ``Are we ready
            for autonomous driving? The KITTI vision benchmark suite,'' in \textit{Proc.
            IEEE Conf. Comput. Vis. Pattern Recognit. (CVPR)}, 2012, pp. 3354--3361.

        \bibitem{menze2015kitti} M. Menze and A. Geiger, ``Object scene flow for
            autonomous vehicles,'' in \textit{Proc. IEEE Conf. Comput. Vis.
            Pattern Recognit. (CVPR)}, 2015, pp. 3061--3070.

        \bibitem{schops2017eth3d} T. Schops, J. L. Schonberger, S. Galliani, T. Sattler,
            K. Schindler, M. Pollefeys, and A. Geiger, ``A multi-view stereo
            benchmark with high-resolution images and multi-camera videos,'' in \textit{Proc.
            IEEE Conf. Comput. Vis. Pattern Recognit. (CVPR)}, 2017, pp. 2538--2547.

        \bibitem{scharstein2014middlebury} D. Scharstein, H. Hirschmuller, Y.
            Kitajima, G. Krathwohl, N. Nesic, X. Wang, and P. Westling, ``High-resolution
            stereo datasets with subpixel-accurate ground truth,'' in \textit{Proc.
            German Conf. Pattern Recognit. (GCPR)}, 2014, pp. 31--42.

        \bibitem{ue5} Epic Games, ``Unreal Engine 5 documentation,'' 2024. [Online].
            Available: \url{https://dev.epicgames.com/documentation/unreal-engine/}

        \bibitem{zedmini} Stereolabs, ``ZED Mini stereo camera,'' 2024.
            [Online]. Available: \url{https://www.stereolabs.com/store/products/zed-mini}
    \end{thebibliography}
    
\end{document}